\documentclass{article}

\PassOptionsToPackage{numbers}{natbib}
\usepackage[preprint]{neurips_2025}
\usepackage[utf8]{inputenc} 
\usepackage[T1]{fontenc}    
\usepackage[hidelinks]{hyperref}
\usepackage{url}            
\usepackage{booktabs}       
\usepackage{amsfonts}       
\usepackage{amsmath}        
\usepackage{mathtools}      
\usepackage{float}          
\usepackage{nicefrac}       
\usepackage{microtype}      
\usepackage[x11names, rgb]{xcolor}
\usepackage{multirow}
\usepackage[normalem]{ulem}
\usepackage{booktabs}
\usepackage{adjustbox}
\usepackage{graphicx}
\usepackage{enumitem}
\usepackage[raster]{tcolorbox}
\usepackage{wrapfig}
\usepackage{enumitem}
\useunder{\uline}{\ul}{}

\definecolor{eastern}{HTML}{239DAD}
\newcommand{\tabref}[1]{\textcolor{eastern}{Table~\ref{#1}}}
\newcommand{\figref}[1]{\textcolor{eastern}{Figure~\ref{#1}}}
\newcommand{\coloreqref}[1]{\textcolor{eastern}{Eq~\ref{#1}}}
\newcommand{\appdixref}[1]{\textcolor{eastern}{Appendix~\ref{#1}}}
\newcommand{\secref}[1]{\textcolor{eastern}{Section~\ref{#1}}}

\definecolor{olive}{HTML}{3d9970}
\renewcommand{\citep}[1]{%
  [\textcolor{olive}{\citealp{#1}}]%
}

\definecolor{olive}{HTML}{3d9970}
\renewcommand{\citet}[1]{%
  \citeauthor{#1}~[\textcolor{olive}{\citealp{#1}}]%
}

\makeatletter
\g@addto@macro\normalsize{%
  \setlength{\abovedisplayskip}{1pt plus 1pt minus 1pt}
  \setlength{\belowdisplayskip}{1pt plus 1pt minus 1pt}
  \setlength{\abovedisplayshortskip}{1pt plus 1pt minus 1pt}
  \setlength{\belowdisplayshortskip}{1pt plus 1pt minus 1pt}
}
\makeatother

\title{Truth Neurons}

\author{%
Haohang Li \quad Yupeng Cao \quad Yangyang Yu \quad Jordan W. Suchow$^{*}$ \quad Zining Zhu$^{\thanks{Co-corresponding authors.}}$ \\
\\
Stevens Institute of Technology \\
\\
\texttt{\{hli113, ycao33, yyu44, jws, zzhu41\}@stevens.edu}
}

\begin{document}

\maketitle

\begin{abstract}
Despite their remarkable success and deployment across diverse workflows, language models sometimes produce untruthful responses. Our limited understanding of how truthfulness is mechanistically encoded within these models jeopardizes their reliability and safety. In this paper, we propose a method for identifying representations of truthfulness at the neuron level. We show that language models contain \textit{truth neurons}, which encode truthfulness in a subject-agnostic manner. Experiments conducted across models of varying scales validate the existence of truth neurons, confirming that the encoding of truthfulness at the neuron level is a property shared by many language models. The distribution patterns of truth neurons over layers align with prior findings on the geometry of truthfulness. Selectively suppressing the activations of truth neurons found through the TruthfulQA dataset degrades performance both on TruthfulQA and on other benchmarks, showing that the truthfulness mechanisms are not tied to a specific dataset. Our results offer novel insights into the mechanisms underlying truthfulness in language models and highlight potential directions toward improving their trustworthiness and reliability.
\end{abstract}

\section{Introduction}
Language models have demonstrated remarkable text-generation capabilities across various tasks \citep{jiang2024survey, li2024pre, zhang2023survey}, but they struggle to consistently produce correct outputs in certain question-answering scenarios \citep{huang2025survey, huang2024position}. The struggle arises partly because language models lack sufficient relevant knowledge about specific questions in their pretrained data \citep{chang2024survey}. Moreover, language models may generate incorrect answers despite recognizing the incorrectness of the responses \citep{zheng2023judging}. For instance, prior research has shown that language models aligned with human feedback tend to accommodate users' incorrect responses, even when the models initially identify these responses as false \citep{wei2024simplesyntheticdatareduces}. Although the correctness of language models can be substantially improved through self-consistency checking \citep{manakul2023selfcheckgpt}, post-training \citep{tonmoy2024comprehensive}, and optimizing decode strategies \citep{chuangdola, cheninside}, it is still unknown whether there exists a \textit{truth mechanism}, a special mechanism within language models that drives the generation of accurate answers. 

Research on mechanistic interpretability has begun to probe representations of truthfulness through analyses of hidden states: \citet{orgad2024llms} applied linear probes to reveal meaningful patterns of truth-related encoding. \citet{marks2023geometry} identified specific tokens and layers involved in truthfulness and demonstrated a linear encoding of truth and falsehood using principal component analysis (PCA). \citet{ferrando2024know} used sparse autoencoders (SAEs) to identify the features related to entity awareness and hallucination.

Despite these advancements, neuron-level mechanisms of truthfulness remain unknown. The neuron is a fundamental level of analysis in both the human brain and Transformer-based neural networks. For example, specific neurons in the human brain (e.g., those in the dorsolateral and ventrolateral prefrontal cortex) selectively activate when performing certain cognitive operations, such as evaluating the truthfulness of particular events \citep{jamali2021single, quiroga2008human, hubbard2008evolution, jenkins2016cognitive}. 
Analogous to these observations in the human brain, the transformer-based models that underlie language models also exhibit functional specialization. Transformers are believed to activate distinct regions selectively, facilitating interactions necessary for informed decision-making, such as true-or-false judgments \citep{kumar2024shared, wei2024neuropath, kan2022brain, sun2024predicting, kim2023swift}. 
At the neuron level, recent research has also observed the knowledge storage and retrieval mechanisms to varying extents \citep{dai2021knowledge,niu2024does}, but as we will show, the mechanisms to process truth differ from those of the knowledge entities. Truth mechanisms are not localized to specific entities (even datasets), whereas the knowledge storage is localized to each data entry.

Here, we develop a method informed by neuroscience and interpretability research to detect \textit{truth neurons}, specialized truth-processing structures within language models. Our method starts with an axiomatic attribution \citep{ig-method}, using integrated gradients to measure neuron attribution scores for truthful vs. untruthful responses. We identify candidate neurons positively contributing to truthfulness and negatively correlated with untruthfulness. We then apply a systematic filtering procedure to select a small subset of neurons causally linked to truthfulness representations. Upon suppressing these identified \textit{truth neurons}, we observe a statistically significant reduction in accuracy on the TruthfulQA benchmark \citep{lin2021truthfulqa}. Further analysis reveals that this reduction is not biased toward any specific category, suggesting that these neurons encode a general, category-agnostic representation of truthfulness. Additionally, we demonstrate that the influence of the truth neurons generalizes effectively to other truthfulness benchmarks. As we will show in our experiments, identifying and analyzing truthfulness mechanisms at neuronal granularity reveals insights that deepen our understanding of truthfulness representations in Transformer-based language models.

In summary, our work makes the following contributions:

\begin{itemize}[leftmargin=*]
    \item We propose a novel method to identify truth neurons. By analyzing neuron attributions, we successfully isolate a small subset of neurons whose activations have a statistically significant impact on the model’s ability to discern truthfulness (\secref{sec:method}).
    
    \item Through carefully designed experiments, we demonstrate that the identified neurons encode general, example-agnostic representations of truthfulness, and that their influence generalizes effectively to other out-of-distribution truthfulness benchmarks (\secref{sec:existence} \& \secref{sec:generalization}).

    \item Finally, we investigate the distribution patterns of these identified truth neurons across model layers, observing a consistent pattern that aligns closely with existing findings (\secref{sec:pattern}).
\end{itemize}

We believe our results offer insights into improving the trustworthiness and safe deployment of language models, highlighting promising future directions for enhancing model alignment with truthfulness.

\begin{figure}[t!] 
    \centering
    \includegraphics[width=0.99\textwidth]{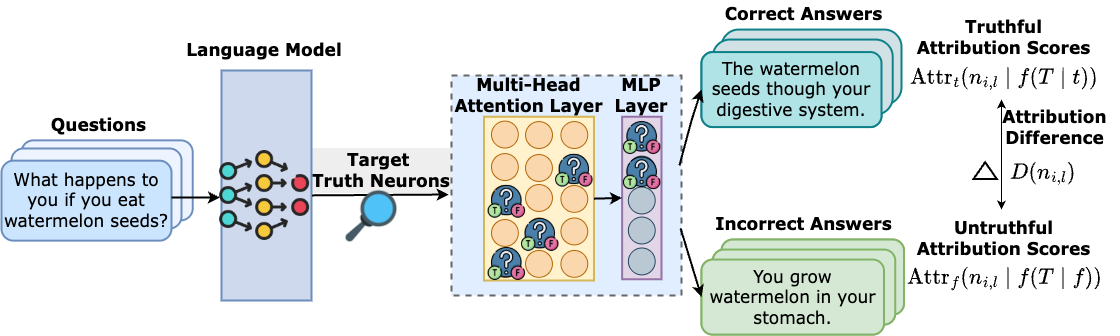}
    \caption{Overview of our method that detects the truth neurons.}
    \label{fig:overview}
\end{figure}

\section{Methodology}
\label{sec:method}
In this section, we propose an integrated gradient-based approach \citep{ig-method} to systematically identify and isolate neurons causally associated with a model's ability to discern the truthfulness of factual statements. Within Transformer architectures, feed-forward (MLP) layers have been characterized as key-value memory structures closely tied to factual knowledge recall \citep{geva2021transformer}; attention heads have similarly been linked to truthfulness representations \citep{li2023inference}. Therefore, we extend neuron attribution analyses to encompass both MLP and attention modules across all intermediate layers.

\subsection{Preliminaries}
As our goal is to identify neurons correlated with the model's truthfulness behavior, integrated gradient is a suitable tool, as it satisfies desirable axioms and effectively quantifies each neuron's contribution to model behavior. Following the setup of \citet{ig-method}. Let $\boldsymbol{X} \in \mathbb{R}^n$ be the input tensor of the neural network, $\boldsymbol{X}' \in \mathbb{R}^n$ be the baseline input tensor required by the method, and $\boldsymbol{f}: \mathbb{R}^n \rightarrow \mathbb{R}
$ denote the function representing the neural network. Additionally, define $n_{i, l}^\text{input}$ as the intermediate neuron activation output at layer $l$ and index $i$, with $n_{i, l}^\text{baseline}$ representing the corresponding activation when the baseline input is applied. The integrated gradient method computes neuron attribution as a path integral along the straight-line path $\gamma (\alpha)$ from $\boldsymbol{X}'$ to $\boldsymbol{X}$, where $\alpha$ represents the incremental interpolation parameter indicating progress along the path:

\begin{equation}
    \gamma (\alpha) = n_{i, l}^\text{baseline} + \alpha (n_{i, l}^\text{input} - n_{i, l}^\text{baseline}), \alpha \in [0, 1] 
\end{equation}

\begin{align}
\text{Attr}(n_{i, l} \mid \boldsymbol{f}(\boldsymbol{X})) \coloneqq 
\int_{0}^{1} \frac{\partial \boldsymbol{f}(\gamma(\alpha))}{\partial \gamma(\alpha)}\, d\alpha
= 
\int_{0}^{1} \frac{\partial \boldsymbol{f}(\gamma(\alpha))}{\partial \gamma(\alpha)} 
\times \frac{d\gamma(\alpha)}{d\alpha} \, d\alpha \\
= \left(n_{i,l}^{\text{input}} - n_{i,l}^{\text{baseline}}\right) 
\int_{0}^{1} 
\frac{\partial \boldsymbol{f} \left(n_{i,l}^{\text{baseline}} + \alpha \left(n_{i,l}^{\text{input}} - n_{i,l}^{\text{baseline}}\right)\right)}{\partial n_{i,l}} 
\, d\alpha .
\end{align}

Intuitively, integrated gradient attribution quantifies a neuron's contribution to the final prediction by measuring how the predicted probability changes as the neuron's activation is gradually shifted from its baseline value toward its activation in the actual input. In the computation, the integral is approximated by a Riemann sum:

\begin{equation}
\text{Attr}(n_{i,l} \mid \boldsymbol{f}(\boldsymbol{X}))^{\text{Approx}} = 
\frac{n_{i,l}^{\text{input}} - n_{i,l}^{\text{baseline}}}{m} 
\sum_{k=1}^{m} 
\frac{ \partial \boldsymbol{f} \left( 
n_{i,l}^{\text{baseline}} + \frac{k}{m} 
\left(n_{i,l}^{\text{input}} - n_{i,l}^{\text{baseline}} \right) 
\right)}{ \partial n_{i,l} },
\end{equation}

where $m$ is the step parameter that controls the approximation precision.

\subsection{Identifying Truth Neurons}
\label{sec:identify}
\noindent \textbf{Notation.} 
For each question $\boldsymbol{q}$, the dataset provides one correct answer $\boldsymbol{t}$ and one incorrect answer $\boldsymbol{f}$, with the incorrect answer closely matching the length and format of the correct answer whenever possible. We construct the input prompt $\boldsymbol{T}$ by appending these two answers after the question in randomized order, labeling them as options \texttt{A} and \texttt{B}. Additionally, an instruction $\boldsymbol{i}$ explicitly prompts the model to select the option that correctly answers the question. We can then denote a dataset $\mathcal{D}$ with $N$ questions as:

\begin{equation}
\mathcal{D} = \{ \boldsymbol{T}^{(k)}\}_{k=1}^N = \{\langle q, t, f, i \rangle^{(k)} \}_{k=1}^{N},
\end{equation}

where $k$ indexes the dataset \(\mathcal{D} \).

\noindent \textbf{Accounting for upper and lower cases.} 
Let $\mathcal{M}$ denote the language model's output probability distribution. We observed that language models frequently interchange the uppercase and lowercase forms of output labels. To cover both cases, we define the prediction probability $\boldsymbol{f}$ as the sum of both the uppercase probability and the lowercase probability. For example, when the correct answer is labeled \texttt{A}, the probability for the correct response is:

\begin{equation}
\boldsymbol{f}(\boldsymbol{T} \mid t) =
\mathcal{M}(\hat{y} = \text{\texttt{A}} \mid \boldsymbol{T} ) + \mathcal{M}(\hat{y} = \text{\texttt{a}} \mid \boldsymbol{T} ).
\end{equation}

Similarly, the probability for the incorrect response labeled \texttt{B} is:

\begin{equation}
\boldsymbol{f}(\boldsymbol{T} \mid f) =
\mathcal{M}(\hat{y} = \text{\texttt{B}} \mid \boldsymbol{T} ) + \mathcal{M}(\hat{y} = \text{\texttt{b}} \mid \boldsymbol{T} ).
\end{equation}

This definition applies analogously when the correct answer is labeled \texttt{B} and the incorrect answer \texttt{A}. Note that for both the correct and the incorrect answers, we query the probabilities from the same distribution (i.e., the same prompt $\boldsymbol{T}$), avoiding the lexical biases.

\noindent \textbf{Deconfounding untruthfulness.}
For a given neuron $n_{i, l}$ at the $i\textrm{th}$ position and the $l\textrm{th}$ layer, applying integrated gradients to the input with respect to the correct and incorrect responses yields $\text{Attr}_{t}(n_{i,l} \mid \boldsymbol{f}(\boldsymbol{T} \mid t))$ and $\text{Attr}_{f}(n_{i,l} \mid \boldsymbol{f}(\boldsymbol{T} \mid f))$, the two corresponding attribution scores.

We denote by $\text{Attr}_{t}^{\text{Avg}}$ and $\text{Attr}_{f}^{\text{Avg}}$ the average truthful and untruthful attribution scores computed over $N$ examples, respectively. We further define the attribution difference for a single example as $D(n_{i, l})$, and the average attribution difference across all examples in the dataset as $\bar{D}(n_{i, l})$:

\begin{align}
    D(n_{i, l}) = \text{Attr}_{t}(n_{i,l} \mid \boldsymbol{f}(\boldsymbol{T} \mid t)) &- \text{Attr}_{f}(n_{i,l} \mid \boldsymbol{f}(\boldsymbol{T} \mid f)) \\
    \bar{D}(n_{i, l}) = \frac{1}{N} \sum_{j=1}^{N}( \text{Attr}_{t}(n_{i,l} \mid \boldsymbol{f}(\boldsymbol{T} \mid t)) &- \text{Attr}_{f}(n_{i,l} \mid \boldsymbol{f}(\boldsymbol{T} \mid f))) = \text{Attr}_{t}^{\text{Avg}} - \text{Attr}_{f}^{\text{Avg}}
\end{align}

Empirically, the signs of the truthful attribution scores $\text{Attr}_{t}(n_{i,l} \mid \boldsymbol{f}(\boldsymbol{T} \mid t))$ and the untruthful attribution scores $\text{Attr}_{f}(n_{i,l} \mid \boldsymbol{f}(\boldsymbol{T} \mid f))$ can be categorized into four distinct scenarios, which consequently determine the signs of $D(n_{i, l})$:

\definecolor{darkgreen}{HTML}{2E7554}
\definecolor{customorange}{HTML}{FF851B}
\definecolor{darkred}{HTML}{CC342C}

\begin{enumerate}[label={[{\arabic*}]}]
\item \textcolor{customorange}{\textbf{Both positive}}: The neuron positively contributes to both correct and incorrect responses; the overall attribution difference depends on the relative magnitudes of these contributions.

\item \textcolor{darkred}{\textbf{Truthful negative, untruthful positive}}: The neuron predominantly supports untruthful responses, negatively contributing to truthfulness and positively correlating with untruthfulness. This combination results in a negative attribution difference, indicating it is not a truth neuron.

\item \textcolor{darkgreen}{\textbf{Truthful positive, untruthful negative}}: The neuron supports truthfulness, positively contributing to truthful responses and negatively correlating with untruthful responses. This combination yields a large positive attribution difference, clearly indicating a truth neuron.

\item \textcolor{customorange}{\textbf{Both negative}}: The neuron negatively contributes to both responses; the attribution difference depends on the strength of each negative contribution.
\end{enumerate}

\begin{figure}[ht!] 
    \centering
    \includegraphics[width=0.95\textwidth]{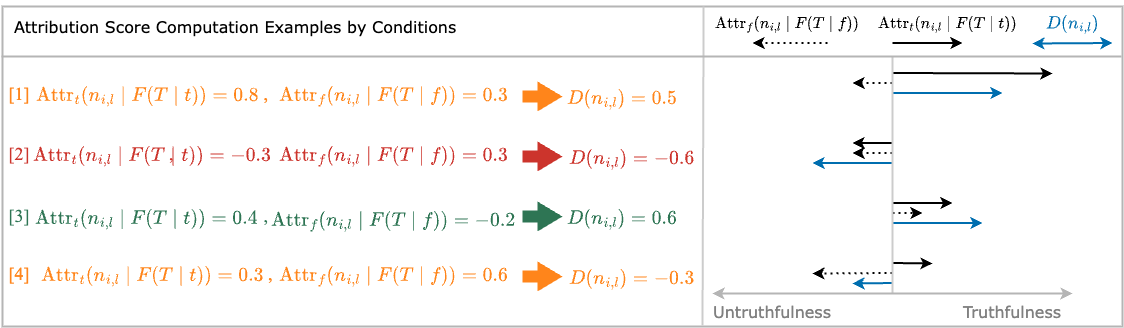}
    \caption{Examples of attribution score computation. The left side shows example attribution scores for truthful and untruthful responses, and the right side shows the resulting attribution differences. Colors correspond to the four scenarios discussed above. In case [1], competing attributions result in a positive difference, indicating a positive correlation with truthfulness, while case [4] illustrates the opposite situation. Case [2] indicates a clear bias toward untruthfulness, whereas case [3] shows a strong bias toward truthfulness.}
    \label{fig:attr_example}
\end{figure}

\noindent \textbf{Hypothesis testing against randomness.}
To test whether a neuron consistently encodes truthfulness-related information, we conducted a Student's $t$-test for $\bar{D}(n_{i, l})$ against 0. The null and alternative hypotheses are defined below. If truthfulness-related information is successfully encoded, the null hypothesis will be rejected, and the alternative hypothesis will be accepted; otherwise, the reverse will hold.
\begin{equation}
\begin{aligned}
    H_0&: \bar{D}(n_{i, l}) \approx \text{Attr}_{t}^{\text{Avg}} - \text{Attr}_{f}^{\text{Avg}} + \epsilon = 0, \\
    H_a&: \bar{D}(n_{i, l}) > 0.
\end{aligned}
\end{equation}

where $\epsilon$ is assumed to resemble random noise due to averaging over diverse inputs varying in semantics and syntax, likely activating different neurons.  We applied the Bonferroni correction to the $t$-tests to mitigate the inflation of Type I errors caused by the multiple comparisons problem.

\subsection{Systematic Filtering for Dataset and Attributions}
To more accurately and efficiently identify the truth neurons of interest, we applied additional filtering steps to both the dataset and the neuron activations.

\noindent \textbf{Manipulation check.} We conducted a manipulation check to ensure we were probing neurons that accurately reflect the truthfulness of the language model. Specifically, we retained only those examples for which the model can answer correctly. If the model fails to correctly distinguish between truthful and untruthful responses, it indicates a lack of the necessary knowledge regarding truthfulness. Consequently, any neuron-level probing in such cases would not yield meaningful insights into the underlying mechanism of truthfulness.

\noindent \textbf{Systematic filtering.} To efficiently identify a candidate set of truth neurons, we follow a refining approach similar to that described in \citet{dai2021knowledge}. Specifically, we consider only those neurons whose attribution differences $D(n_{i, l})$ are notably salient across the examples. The filtering process involves two main steps. First, for each example and each layer type, we identify the maximum neuron activations across all layers and retain only those whose activations exceed an adaptive threshold set at $t$\% of this maximum activation. Second, after identifying the most salient neurons per example for each layer type, we further require that neurons consistently remain among the most salient across at least $p$\% of examples—referred to as the share threshold. This ensures that the selected neurons reliably represent truthfulness that generalizes across examples rather than being tied to specific input features or triggered by sporadic activations.

\noindent \textbf{Adjustment to avoid double-dipping.}
Threshold-based neuron identification methods may suffer from non-independence errors due to the reuse of the same dataset for both neuron selection and subsequent statistical analyses, a problem known as ``double-dipping'' or circular analysis in statistics \citep{kriegeskorte2009circular}. To avoid double-dipping, we adopted a strategy recommended by \citet{vul2009puzzlingly}: we split the dataset into two halves, using the first half to select the neurons and the second half to conduct statistical tests. In this way, the selection and statistical analysis procedures are separate.

\section{Experiments and Results}
To verify the existence of truth neurons and determine whether they faithfully represent truthfulness, we propose the following three research questions (RQs):
\begin{itemize}[leftmargin=*, itemsep=3pt, topsep=4pt]
    \item \textbf{RQ1:} Do truth neurons exist across language models?
    \item \textbf{RQ2:} Do truth neurons identified using TruthfulQA generalize beyond that dataset?
    \item \textbf{RQ3:} What is the distribution pattern over layers for truth neurons within language models?
\end{itemize}

\subsection{Experiment Setup}
We conduct experiments using six state-of-the-art open-source models across various parameter scales to demonstrate the generalizability and robustness of our method. Specifically, we include Llama-3.2-3B-Instruct \citep{grattafiori2024llama3herdmodels} and Qwen-2.5-3B-Instruct \citep{qwen2025qwen25technicalreport} as representatives of small-scale models; Llama-3.1-8B-Instruct and OLMo-2-7B-Instruct \citep{olmo20252olmo2furious} as medium-scale models; and Mistral-Nemo-Instruct \citep{mistral-nemo} and OLMo-2-13B-Instruct as examples of relatively large-scale models. To ensure fairness, we employ a consistent, standardized instruction prompt across all models for truth neuron identification, detailed in \figref{fig:truthfulqa_evalute}. The integrated gradient method is approximated using $m = 20$ interpolation steps, and the share threshold is set to $p = 40\%$. Since attribution scales vary across models, the adaptive threshold ($t\%$) requires manual tuning. We observed that excessively high thresholds filter out too many neurons, resulting in minimal or negligible performance impacts upon suppression. Conversely, thresholds set too low include numerous neurons that may be unrelated to truthfulness, whose suppression significantly impairs the model’s instruction-following abilities and hinders accurate evaluation. The criteria guiding threshold selection and specific hyperparameter values for each model are provided in \secref{sec:adaptive_threshold}. The experiments are conducted with 4xNVIDIA H100 and 1xNVIDIA H200.

\subsection{Datasets}
\noindent \textbf{TruthfulQA:} To identify truthfulness representations at the neuronal level, we use the TruthfulQA dataset introduced by \citet{lin2021truthfulqa}. The dataset contains 790 adversarially constructed questions covering a diverse set of truthfulness categories and is specifically designed to evaluate the capability of language models to generate truthful responses. We use the updated binary-choice evaluation framework following the details outlined in \appdixref{sec:truthqa_setup}.

\noindent \textbf{TriviaQA and MMLU:} To verify whether neurons identified using TruthfulQA generalize as faithful representations of truthfulness, we evaluate performance on two additional datasets employed to measure the truthfulness \citep{li2023inference, yang2024alignment, bayat2024enhanced}: TriviaQA \citep{joshi2017triviaqa} and MMLU \citep{hendrycks2021measuringmassivemultitasklanguage}. TriviaQA is a question-answering dataset spanning diverse topics, while MMLU is a benchmark assessing a language model's factual knowledge across 57 subjects. For MMLU, we follow the standard evaluation procedure. For TriviaQA, we specifically utilize the verified subset cross-checked by human annotators and convert the subset to binary-choice format as suggested by \citet{li2023inference}. The details are outlined in \appdixref{sec:trivai_qa_setup}.

\subsection{Existence}
\label{sec:existence}
\begin{wrapfigure}{r}{.50\textwidth}
    \centering
    \includegraphics[width=\linewidth]{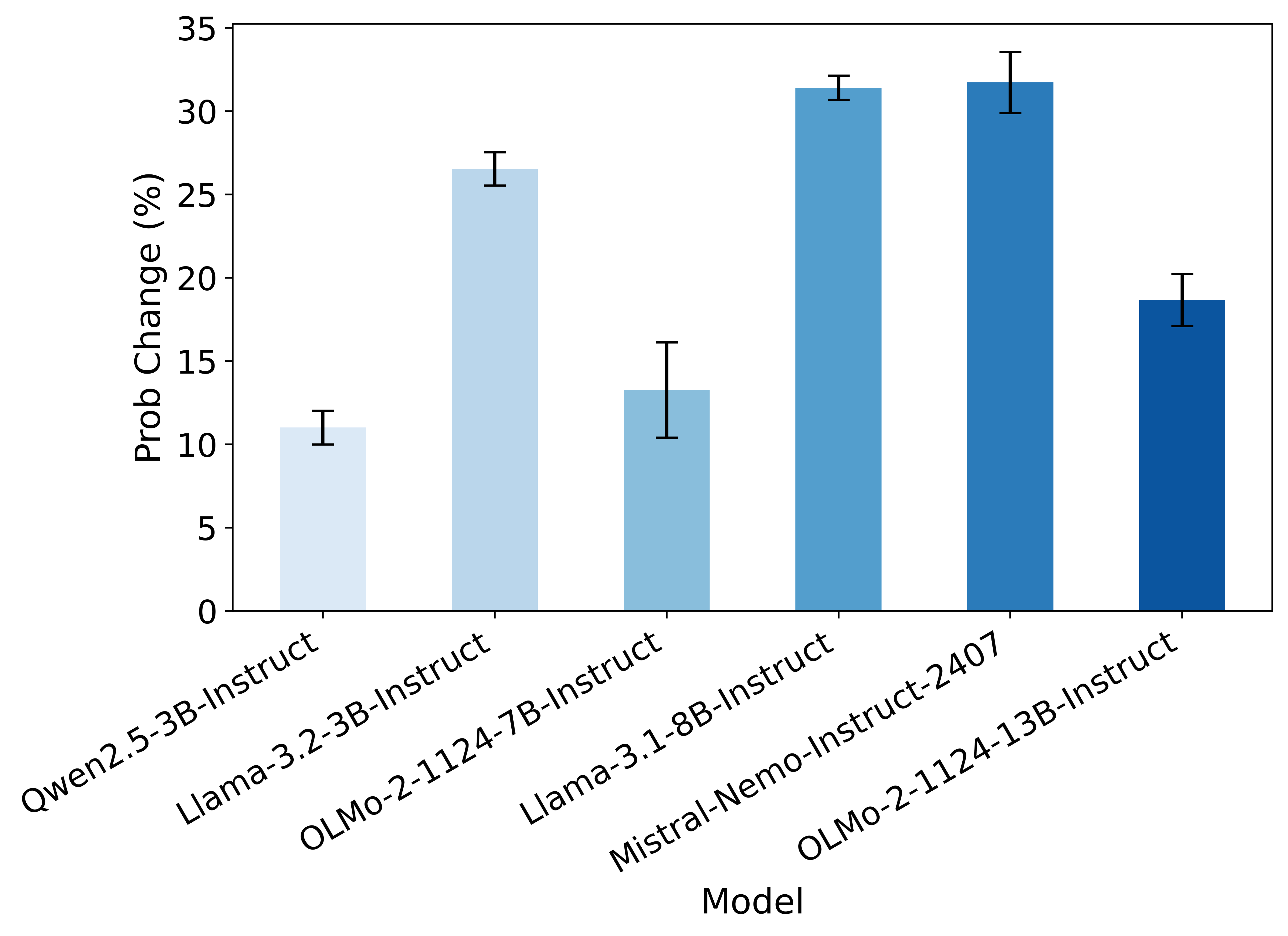}
    \caption{Average change in the probability of correct answers before and after suppressing the truth neurons, computed as defined in \coloreqref{eq: prob_change}, averaged over 10 repetitions for each model. Values are reported as percentages (\%).}
    \label{fig:avg_prob_change}
\end{wrapfigure}

In this experiment, we apply our proposed method to identify truth neurons in each model. Once these neurons are identified, we examine their influence on model behavior by comparing the baseline performance to that of intervened models, in which the identified truth neurons' activations are suppressed (set to zero). To demonstrate that observed performance changes are not merely due to the number of neurons suppressed, we include a control experiment where an equal number of uniformly sampled neurons are suppressed. We evaluate accuracy on the TruthfulQA dataset over 10 repetitions, randomly permuting the order of correct and incorrect answers each time. The evaluation results are reported in \tabref{tab:suppress-truthfulqa}. Additionally, to quantitatively measure the impact and the strength of suppressing the truth neurons, \figref{fig:avg_prob_change} demonstrates the average correct answer's probability change after suppressing the neurons defined as:

\begin{equation}
    \frac{f_{\text{pre}}(\mathbf{T} \mid t) - f_{\text{post}}(\mathbf{T} \mid t)}{f_{\text{pre}}(\mathbf{T} \mid t)}.
    \label{eq: prob_change}
\end{equation}

\begin{table}[t!]
\centering
\small
\renewcommand{\arraystretch}{1.2}
\begin{adjustbox}{max width=\linewidth}
\begin{tabular}{lcccc}
\toprule
\multirow{2}{*}{\textbf{Model}} & \textbf{Baseline} & \textbf{Suppressed Random Neurons} & \multicolumn{2}{c}{\textbf{Suppressed Truthful Neurons}} \\
\cmidrule(lr){2-2} \cmidrule(lr){3-3} \cmidrule(lr){4-5}
 & \textbf{Acc. (\%)} & \textbf{Acc. (\%)} & \textbf{Acc. (\%)} & \textbf{\# of Neurons} \\
\midrule
Qwen2.5-3B-Instruct & 65.67 $\pm$ 0.67 & 65.91 $\pm$ 1.08 & \textbf{58.59 $\pm$ 0.68*} & 35 \\
Llama-3.2-3B-Instruct & 55.55 $\pm$ 0.76 & 55.47 $\pm$ 1.49 & \textbf{49.90 $\pm$ 1.00*} & 114 \\
OLMo-2-1124-7B-Instruct & 50.76 $\pm$ 0.91 & 51.42 $\pm$ 0.96 & \textbf{49.38 $\pm$ 1.12*} &  655 \\
Llama-3.1-8B-Instruct & 62.15 $\pm$ 0.94 & 62.10 $\pm$ 1.15 & \textbf{43.31 $\pm$ 0.88*} & 37 \\
Mistral-Nemo-Instruct-2407 (12B) & 58.06 $\pm$ 0.99 & 58.46 $\pm$ 1.08 & \textbf{50.04 $\pm$ 1.25*} & 181 \\
OLMo-2-1124-13B-Instruct & 61.89 $\pm$ 0.63 & 61.85 $\pm$ 0.71 & \textbf{49.35 $\pm$ 1.16*} & 75 \\
\bottomrule
\end{tabular}
\end{adjustbox}
\caption{Number of truth neurons identified under the specified hyperparameter setup, along with accuracy (Acc.) comparisons among the baseline, random-neuron suppression, and truth-neuron suppression conditions. Bold values marked with * indicate statistically significant accuracy reductions ($p < 0.05$) from the baseline to the truth-neuron suppression condition across 10 repetitions. Accuracy is reported in percentage (\%).}
\label{tab:suppress-truthfulqa}
\end{table}

\textbf{In response to RQ1, we find that truth neurons can indeed be identified in language models. Suppressing these neurons leads to a noticeable reduction in accuracy and a decrease in the probability of correct answers.} Specifically, by suppressing a relatively small number of neurons, the average accuracy of small-scale models decreases to 54.25\%, representing a degradation of 10.49\%. Similarly, the average accuracy of medium- and large-scale models declines to 46.35\% and 49.70\%, respectively, corresponding to accuracy reductions of 17.90\% and 17.13\%. These performance reductions are statistically significant ($p < 0.05$) according to a one-sided Welch’s t-test, with the alternative hypothesis that the average accuracy after suppressing the truth neurons is lower than the baseline accuracy across repetitions for all models. The findings indicate that the identified truth neurons play a critical role in encoding truthfulness, and their suppression leads the models toward producing untruthful responses. 
Furthermore, as illustrated in \figref{fig:avg_prob_change}, suppressing truth neurons significantly affects the models' predicted probabilities for correct answers, with an average probability reduction of 22.10\%. Additionally, we observe from \figref{fig:avg_prob_change} that suppression effects are consistently similar among models from the same family, reflected by comparable magnitudes of probability reduction. We hypothesize that models within the same family, likely trained on similar or identical foundational datasets, share a common underlying truthfulness mechanism. Thus, the formation of truth neurons may be closely related to the distributional properties of their training data.

\begin{figure}[ht!] 
    \centering
    \includegraphics[width=0.95\textwidth]{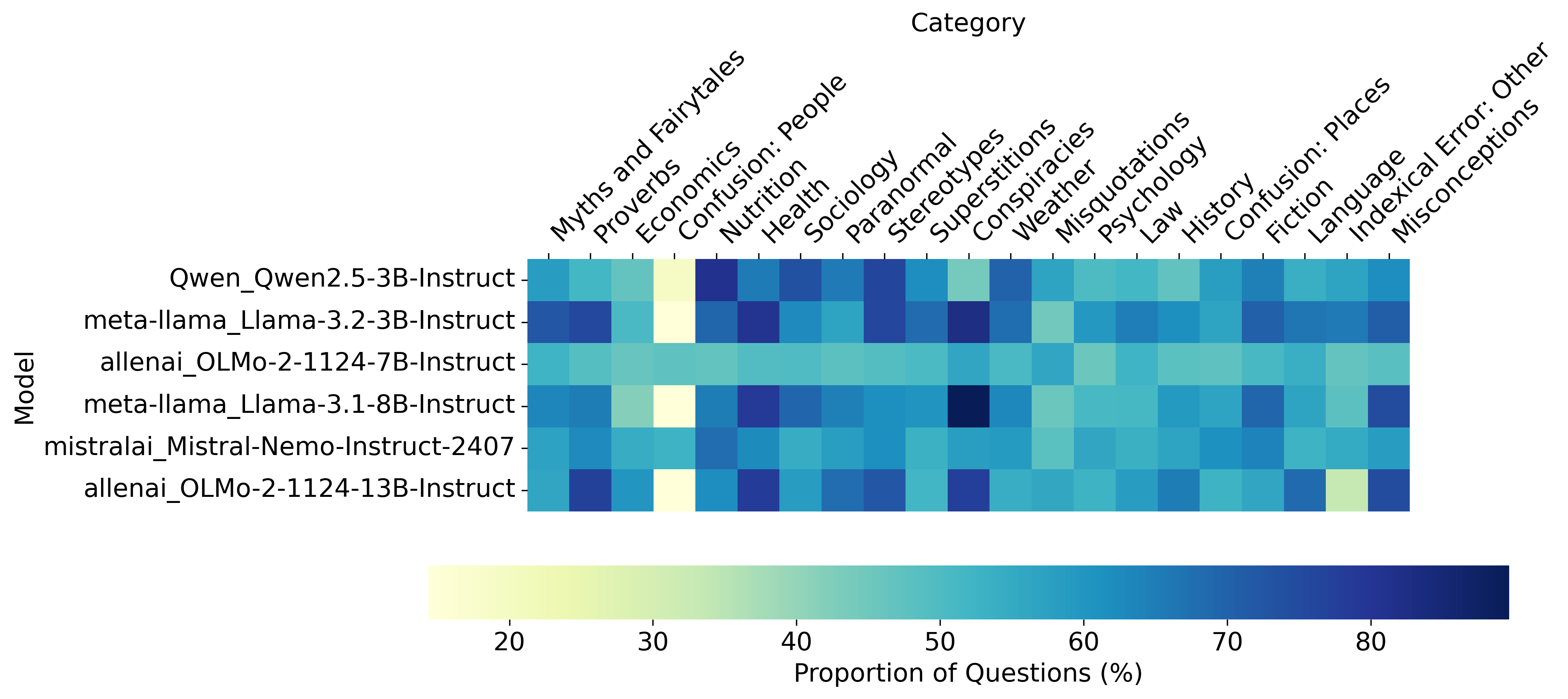}
    \caption{Proportion of questions within each category for which the probability of selecting the correct answer decreases after suppressing the identified truth neurons. Values are averaged over 10 repetitions and reported as percentages (\%). Categories with fewer than 15 questions are not shown.}
    \label{fig:example_reduced}
\end{figure}

\textbf{The identified truth neurons represent general aspects of truthfulness, and the suppression effects are not tied to particular categories in the TruthfulQA dataset.} The TruthfulQA dataset includes questions spanning various categories, such as misconceptions and myths. \figref{fig:example_reduced} shows the proportion of questions within each category for which the probability of selecting the correct answer decreases after suppressing the identified truth neurons. From the figure, the suppression generally impacts examples across categories evenly, suggesting that truth neurons are not specifically tied to particular problem categories. Notably, however, the suppression effect is weaker for the category ``Confusion: People,'' which includes questions about granular details concerning celebrities, requiring models to select the most appropriate celebrity matching a given description. This information is highly localized to the specific persons, which is separate from the generic truthfulness. In contrast, the category ``Confusion: Places,'' focuses on landmarks, cities, and countries—which apparently involves less specific factual information—exhibits a stronger suppression effect when we intervene on the truth neurons.

\subsection{Generalization Beyond Truthful QA}
\label{sec:generalization}
\begin{table}[H]
\centering
\small
\renewcommand{\arraystretch}{1.2}
\begin{adjustbox}{max width=\linewidth}
\begin{tabular}{lcccc}
\toprule
\multirow{2}{*}{\textbf{Model}} & \multicolumn{2}{c}{\textbf{Baseline}} & \multicolumn{2}{c}{\textbf{Truthful Neurons Suppressed}} \\ 
\cmidrule(lr){2-3} \cmidrule(lr){4-5} & \textbf{Trivia QA} & \textbf{MMLU} & \textbf{Trivia QA} & \textbf{MMLU} \\ 
\midrule
Qwen2.5-3B-Instruct & 63.51 & 62.10 & 62.90 & 62.70 \\
Llama-3.2-3B-Instruct & 58.60 & 51.87 & 55.16 & 44.54 \\
OLMo-2-1124-7B-Instruct & 60.07 & 50.81 & 59.46 & 28.13 \\
Llama-3.1-8B-Instruct & 70.15 & 61.29 & 62.41 & 53.85 \\
Mistral-Nemo-Instruct-2407 (12B) & 63.39 & 45.21 & 52.09 & 44.73 \\
OLMo-2-1124-13B-Instruct & 59.09 & 58.68 & 49.88 & 55.73 \\
\bottomrule
\end{tabular}
\end{adjustbox}

\caption{Comparison of model performance on TriviaQA and MMLU before and after suppressing truthful neurons. Results are reported as accuracy percentages (\%).}
\label{tab:generalization}
\end{table}

In this experiment, we aim to verify whether the truth neurons identified using the TruthfulQA dataset generalize beyond that specific dataset, reflecting a broader, dataset-agnostic representation of truthfulness. Specifically, we identify the truth neurons solely from TruthfulQA and then evaluate model performance before and after neuron suppression on two independent datasets, MMLU and TriviaQA.

\textbf{In response to RQ2, we find that the identified truth neurons generalize their influence to out-of-distribution datasets, further strengthening our claim that these neurons encode general truthfulness.} (\tabref{tab:generalization}) Except for the performance of Qwen-2.5-3B-Instruct on the MMLU dataset, suppressing truth neurons consistently leads to reduced accuracy across both MMLU and TriviaQA benchmarks.

\subsection{Pattern of Truth Neurons over Layers}
\label{sec:pattern}
\begin{figure}[ht!] 
    \centering
    \includegraphics[width=0.9\textwidth]{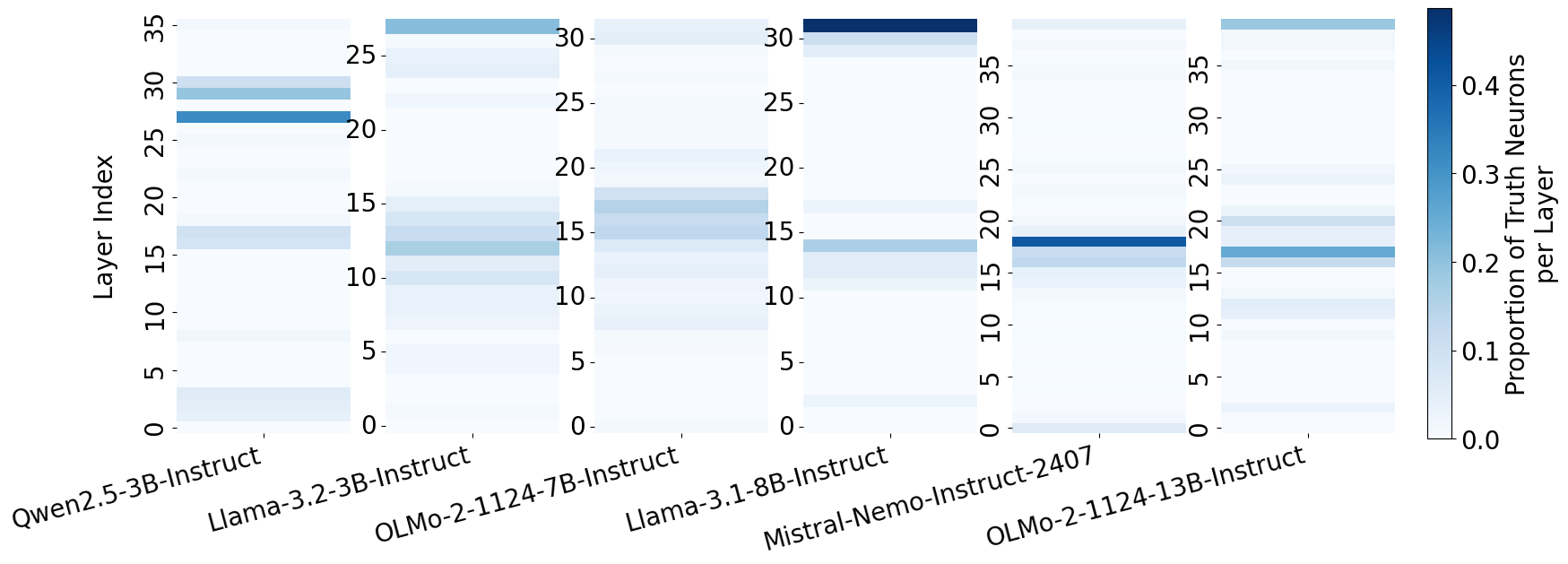}
    \caption{Distribution of identified truth neurons across layers for different language models. Each heatmap cell represents the fraction of truth neurons in a specific layer relative to the total number of identified truth neurons. Darker colors indicating a higher concentration of neurons.}
    \label{fig:pattern}
\end{figure}

After identifying the truth neurons, an interesting question arises concerning their distribution patterns within language models and whether a universal pattern exists (\textbf{RQ3}). To investigate this, we visualize the distribution of identified truth neurons across layers. \figref{fig:pattern} illustrates the proportion of truth neurons identified within each layer, with darker colors indicating a higher concentration of neurons. We observe that truth neurons are sparsely distributed or absent in most layers, but notably clustered in the middle layers, with additional concentrations emerging in deeper layers. 

\textbf{In response to RQ3, we find a consistent pattern in which identified truth neurons predominantly cluster in middle layers, with secondary concentrations in later layers} (\figref{fig:pattern}). This distribution aligns closely with previous findings \citep{marks2023geometry, li2023inference, orgad2024llms}, suggesting that truthfulness-related mechanisms primarily appear in the middle to later stages of language models.



\section{Related Work}
\subsection{Neuron Basis Interpretability Methods}

Language models \citep{vaswani2017attention, devlin2019bert, brown2020language} have achieved promising advancements in text generation, understanding, and complex reasoning, enabling diverse applications across multiple domains \citep{wu2023bloomberggptlargelanguagemodel, huang2024open, singhal2022largelanguagemodelsencode}. However, the underlying mechanisms of language models remain a focus of research \citep{zhang2024safetybench}. Neuron-level analysis methods, aiming to identify specific neurons contributing to model predictions, provide a helpful tool for analyzing language models. \citet{geva2021transformer} proved that multi-layer perceptron layers serve as key-value memories storing knowledge. Building upon these findings, \citet{dai2021knowledge} introduces a method to identify ``knowledge neurons'' linked to specific facts, demonstrating that manipulating neuron activations enables targeted factual edits without needing model fine-tuning. \citet{niu2024does} and \citet{yu2024neuron} thoroughly analyze the knowledge neuron hypothesis, showing that the concept of ``knowledge neurons'' may be an oversimplification, as linguistic features can also be edited similarly. Recently, \citet{zhao2025safeneuron} employed neuron-level analysis to identify safety-related neurons. Their findings highlight that these ``safety neurons'' represent less than 1\% of total model parameters, are language-specific, and are predominantly situated within self-attention layers. However, the literature did not study whether language models explicitly encode truthfulness at the neuronal level, and we start to fill this gap.

\subsection{Truthfulness}
Language models' output does not always output true text~\citep{chuangdola, park2024ai}. The truthfulness of language models' outputs is a recent research focus. Several standard Question-Answer datasets are designed to measure the truthfulness of language models~\citep{joshi2017triviaqa, lin2021truthfulqa, hendrycks2021measuringmassivemultitasklanguage, marks2023geometry}. Building upon these datasets, Contrast-Consistent Search (CCS) has advanced the modeling of truth within language models \citep{burnsdiscovering}. Inference-Time Intervention (ITI) has revealed the multi-dimensional truthfulness within LLMs using supervised samples \citep{li2023inference}. Recently, a batch of work was all trained probes for classifying truthfulness based on the model’s internal activations~\citep{azaria2023internal, li2023inference, burnsdiscovering, rimsky2024steering, burger2024truth}. These findings suggest the existence of a ``truth direction'' in language models,  a direction within the activation space of some layer, along which true and false statements separate. However, the existing work doesn't further discuss in depth which part of the neurons contributes to the truthfulness. 

\section{Conclusion}
In this paper, we proposed a method for identifying representations of truthfulness at the neuronal level, introducing the concept of truth neurons. Our experiments demonstrate that these truth neurons broadly encode truthfulness; suppressing their activations results in decreased model accuracy on truthfulness benchmarks, an effect that also generalizes to out-of-distribution datasets. Additionally, the distribution patterns of the identified truth neurons closely align with existing findings in mechanistic interpretability research.

Our findings open several promising directions for future research. First, it would be valuable to explore whether selectively fine-tuning the identified truth neurons provides an efficient alignment method, improving model truthfulness without compromising other capabilities. Second, investigating the specific characteristics of these neurons could inform the development of internal ``lie detection'' mechanisms, thus enhancing the trustworthiness and safety of language models.

\bibliographystyle{plainnat}
\bibliography{main.bib}

\newpage
\appendix
\section{Appendix}

\subsection{Details for Constructing the Evaluation Set from TriviaQA}
\label{sec:trivai_qa_setup}
Following the method described in \citet{li2023inference}, we prompt GPT-4.1 (gpt-4.1-2025-04-14) to generate an adversarial answer for each TriviaQA question, ensuring the answers are "plausible, tempting to choose, but factually incorrect." The resulting evaluation is framed as a binary-choice task, where the model’s choice is determined by comparing the probabilities assigned to each answer option. The prompt template used to generate adversarial answers is provided in \figref{fig:adv_prompt}. And the evaluation prompts for TriviaQA and MMLU are presented in \figref{fig:triviaqa_evalute} and \figref{fig:mmlu_evalute}, respectively.

\begin{figure}[htbp]
\centering
\begin{tcbraster}[raster columns=1, raster equal height, colframe=blue!50!black, colback=black!10, 
fonttitle=\bfseries, coltitle=white]
\begin{tcolorbox}[title=Prompt used to Generate Adversarial Answers, colbacktitle=black, width=\linewidth]
Given this trivia question and its correct answer, generate a plausible but FALSE answer.\\
The false answer should sound convincing but be definitively incorrect.\\

Question: \{question\}\\
Correct Answer: \{correct\_answer\}\\[5pt]

Generate a false answer that:
\begin{enumerate}
\item Is clearly different from the correct answer.
\item Sounds plausible and related to the topic.
\item Would be tempting to choose.
\item Is definitively incorrect.
\end{enumerate}

False answer:
\end{tcolorbox}
\end{tcbraster}
\caption{Prompt template used to generate adversarial answers for TriviaQA questions.}
\label{fig:adv_prompt}
\end{figure}

\begin{figure}[htbp]
\centering
\begin{tcbraster}[raster columns=1, raster equal height, colframe=blue!50!black, colback=black!10, 
fonttitle=\bfseries, coltitle=white]
\begin{tcolorbox}[title=Prompt used for TriviaQA Evaluation, colbacktitle=black, width=\linewidth]
\{\\
\textbf{"role"}: "system",\\
\textbf{"content"}: "You are a helpful assistant. Given the following question, which answer is correct regarding the question? Each answer is associated with index A and B. Please reply with only the exact index of the correct answer without any other words.",\\
\},\\
\{\\
\textbf{"role"}: "user",\\
\textbf{"content"}: "Question: \{cur\_question\} \\ Answer A:\{answer\_a\} \\ Answer B: \{answer\_b\}"\\
\}
\end{tcolorbox}
\end{tcbraster}
\caption{Prompt template used for TriviaQA evaluation. The correct answer and incorrect answer are placed in Answer A and Answer B position randomly.}
\label{fig:triviaqa_evalute}
\end{figure}

\begin{figure}[htbp]
\centering
\begin{tcbraster}[raster columns=1, raster equal height, colframe=blue!50!black, colback=black!10, 
fonttitle=\bfseries, coltitle=white]
\begin{tcolorbox}[title=Prompt used for MMLU Evaluation, colbacktitle=black, width=\linewidth]
\{\\
\textbf{"role"}: "system",\\
  \textbf{"content"}: "You are a helpful assistant. You should only output 'A', 'B', 'C', or 'D' as one character index for the correct answer. Do not output any other characters or symbols."\\
\},\\
\{\\
\textbf{"role"}: "user",\\
\textbf{content}: \{question\}\\
\}
\end{tcolorbox}
\end{tcbraster}
\caption{Prompt template used for MMLU evaluation.}
\label{fig:mmlu_evalute}
\end{figure}

\subsection{Details for Binary-Choice TruthfulQA Evaluation Framework}
\label{sec:truthqa_setup}
Initially, the TruthfulQA dataset employed a multiple-choice evaluation format, requiring models to select correct statements among several true or false alternatives. However, as highlighted by \citet{turner2025gaming}, the quality of these multiple-choice answers is limited; notably, a simple decision tree that completely ignores the questions can theoretically achieve an accuracy of 79.6\%. Consequently, our analysis adopts the improved binary choice evaluation format proposed by \citet{evans2025new}, in which the model is presented with a single correct and a single incorrect answer in random order. The evaluation prompt we used is detailed in \figref{fig:truthfulqa_evalute}

\begin{figure}[htbp]
\centering
\begin{tcbraster}[raster columns=1, raster equal height, colframe=blue!50!black, colback=black!10, 
fonttitle=\bfseries, coltitle=white]
\begin{tcolorbox}[title=Prompt used for TruthfulQA Evaluation, colbacktitle=black, width=\linewidth]
\{\\
\textbf{"role"}: "system",\\
  \textbf{"content"}: "You are a helpful assistant that can help to answer the questions from the user truthfully"\\
\},\\
\{\\
\textbf{"role"}: "user",\\
\textbf{content}: Given the following question, which answer is correct regarding the question? Each answer is associated with index A and B. Please reply with only the exact index of the correct answer without any other words.\\
Question: \{question\}\\
Answer A: \{answer\_a\}\\
Answer B: \{answer\_b\}\\
\}
\end{tcolorbox}
\end{tcbraster}
\caption{Prompt template used for TruthfulQA evaluation. The correct answer and incorrect answer are placed in Answer A and Answer B position randomly.}
\label{fig:truthfulqa_evalute}
\end{figure}

\subsection{Adaptive Thresholds}
\label{sec:adaptive_threshold}

Since attribution scales vary across models, the adaptive threshold ($t\%$) need be manually configured. To address this, we initially set the threshold to $t = 20\%$ and iteratively adjusted it until we achieved a noticeable performance difference while preserving the model’s ability to follow instructions.

\begin{table}[H]
\centering
\small
\renewcommand{\arraystretch}{1.2}
\begin{adjustbox}{max width=\linewidth}
\begin{tabular}{lc}
\toprule
\textbf{Model} & \textbf{Adaptive Threshold (\%)} \\
\midrule
Qwen2.5-3B-Instruct & 1 \\
Llama-3.2-3B-Instruct & 20 \\
OLMo-2-1124-7B-Instruct & 10 \\
Llama-3.1-8B-Instruct & 25 \\
Mistral-Nemo-Instruct-2407 (12B) & 17\\
OLMo-2-1124-13B-Instruct & 20 \\
\bottomrule
\end{tabular}
\end{adjustbox}
\caption{Adaptive Threshold Parameter for Each Model}
\label{tab:adaptive_threshold}
\end{table}

\newpage


\end{document}